\documentclass[conference]{IEEEtran}
\usepackage{cite}
\usepackage{amsmath,amssymb,amsfonts}
\usepackage{algorithmic}
\usepackage{graphicx}
\usepackage{textcomp}
\usepackage{xcolor}
\usepackage{subfig}
\def\BibTeX{{\rm B\kern-.05em{\sc i\kern-.025em b}\kern-.08em
    T\kern-.1667em\lower.7ex\hbox{E}\kern-.125emX}}
\begin{document}

\title{3D printed realistic finger vein phantoms}

\author{\IEEEauthorblockN{1\textsuperscript{st} Luuk Spreeuwers}
\IEEEauthorblockA{\textit{Faculty of EEMCS} \\
\textit{University of Twente}\\
Enschede, Netherlands \\
l.j.spreeuwers@utwente.nl}
\and
\IEEEauthorblockN{2\textsuperscript{nd} Rasmus van der Grift}
\IEEEauthorblockA{\textit{Faculty of EEMCS} \\
\textit{University of Twente}\\
Enschede, Netherlands \\
}
\and
\IEEEauthorblockN{3\textsuperscript{rd}  Pesigrihastamadya Normakristagaluh}
\IEEEauthorblockA{\textit{Faculty of EEMCS} \\
\textit{University of Twente}\\
Enschede, Netherlands \\
\IEEEauthorblockA{\textit{Indonesian Institute of Sciences (LIPI)}\\
Indonesia} \\
}
}

\maketitle

\begin{abstract}
Finger vein pattern recognition is an emerging biometric with a good resistance to presentation attacks and low error rates. One problem is that it is hard to obtain ground truth finger vein patterns from live fingers. In this paper we propose an advanced method to create finger vein phantoms using 3D printing where we mimic the optical properties of the various tissues inside the fingers, like bone, veins and soft tissues using different printing materials and parameters. We demonstrate that we are able to create finger phantoms that result in realistic finger vein images and precisely known vein patterns. These phantoms can be used to develop and evaluate finger vein extraction and recognition methods.
In addition, we show that the finger vein phantoms can be used to spoof a finger vein recognition system.\\
This paper is based on the Master's thesis of Rasmus van der Grift.
\end{abstract}


\section{Introduction}
Fingerprint or face recognition is increasingly becoming an integral part of personal devices. Another emerging biometric technology is blood vessel based recognition and identification like e.g. finger vein pattern recognition. Advantages of finger vein pattern based biometric recognition are good resistance to presentation attacks, very low error rates and user convenience comparable to fingerprint recognition \cite{handbook}.
Face and fingerprint recognition systems are often tested and optimized using data with their ground truths. These ground truths are relatively easy to obtain because this information is present on the surface of the body. Since blood vessels are present inside the human tissue, obtaining the ground truth is not as easy. 
In earlier work we proposed synthesised finger vein images \cite{fieke}, that allow us to obtain ground truth vein patterns, however, these images are not as realistic as images obtained using 3D phantoms. In our first attempts to create 3D finger phantoms, we used soap, silicon and wires and 3D printed 'bones' \cite{pesi1,pesi3}. However, construction was complicated and did not allow for complex vein patterns. 
In this paper we show that using 3D printing, a complete phantom of a finger can be created where the properties of all important tissues are mimicked with complex vein structures and where the ground truth is known. This technique can be used to create a data set that can be used to help design finger vein acquisition devices and improve finger vein recognition systems e.g. for cross device finger vein recognition, see \cite{tugce2}. This paper is based on the Master's report of Rasmus van der Grift \cite{rasmus}.

\section{Related work}
\subsection{Acquisition of finger vein patterns}
\begin{figure}[ht]
	\centering
	\includegraphics[width=\columnwidth]{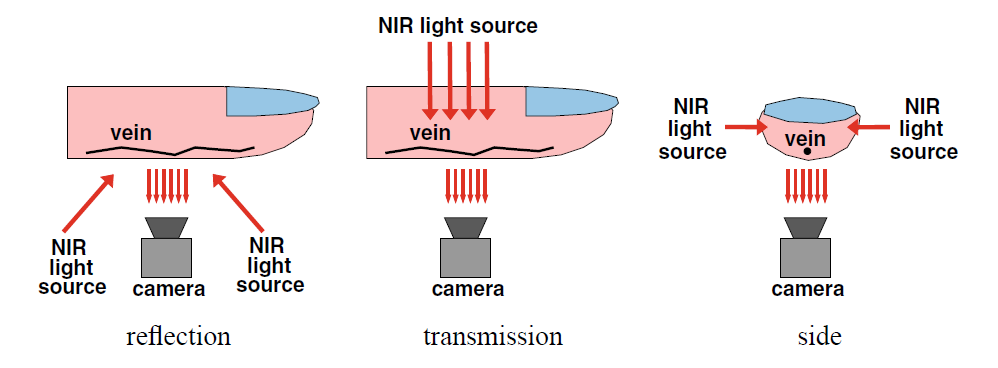}
	\caption{Acquisition of finger vein patterns}
	\label{fig1}
\end{figure}

The ’handbook of Vascular Biometrics’\cite{handbook} describes different types of sensors for finger vein recognition. Because haemoglobin in veins has a higher absorption of Near-Infrared (NIR) light than the surrounding tissue, the vascular pattern inside a finger can be captured by a device sensitive to NIR light. There are several ways to illuminate the finger to extract blood vessels. The main types found in existing devices are shown in Figure \ref{fig1}.

\begin{figure}[ht]
	\centering
	\includegraphics[width=0.8\columnwidth]{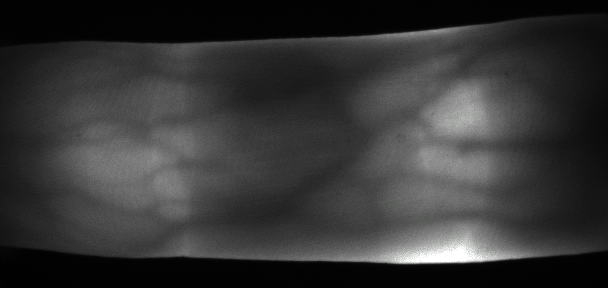}
	\caption{NIR image of a finger with veins}
	\label{fig22}
\end{figure}

Irrespective of the type of illumination, the veins will be visible as dark, rather unsharp lines against a brighter background, see Figure \ref{fig22}. A more in-depth analysis of the imaging process and the optical behaviour of the tissues in the finger is presented in \cite{pesi1,muriel}.

From these studies, we know that the bone acts as a diffuser, scattering the light in all directions, while it only absorbs part of the NIR light. The softer tissues (fat etc.) absorb the NIR more than the bone and the blood in the veins absorbs the NIR even more. Because the bone near the joints is thicker there is less soft tissue and, consequently, the image appears brighter at the joints.

In this paper we used the finger vein scanning device that was developed at the University of Twente and provides high quality vein images \cite{bram1,bram2}.

\subsection{Depth of bloodvessels}
The contrast, size and sharpness of the depicted bloodvessels depends on their size and depth, i.e. distance to the skin. This is described in \cite{cupera}. If the bloodvessels are further away from the skin, their contrast and sharpness decreases. The veins at the joints are closest to the surface and, hence, generally more sharply visible. So at the joints, the image is both brighter and the veins are sharper. This can be observed in Figure \ref{fig22}.

\subsection{Finger vein recognition}
A biometric system for identifying individuals by the pattern of veins in a finger using the maximum curvature was proposed by Miura \cite{miura}. It is based on simple correlation of binary vein patterns. Even though many other methods for vein recognition have been proposed, among others based on deep learning networks, see e.g \cite{tugce}, the method proposed by Miura performs not much worse than more modern methods and is still used as a good baseline.
An important step
in finger vein recognition is proper alignment of
the finger vein patterns. This alignment method
is used to deal with variations in finger pose. In \cite{pesi2} we proposed an ICP alignment approach that improves the recognition results considerably.

\section{Methods}
In order to create realistic finger vein phantoms using 3D printing, the following steps were taken:

\begin{itemize}
	\item Select PLA printing material with optical properties similar to the tissues in the finger
	\item Define shape of 3D model of bones and soft tissues
	\item Define 3D vein patterns
	\item Combine into complete finger phantom and create 3D model for printing
\end{itemize}
The different steps are detailed in the sections below.

\subsection{PLA 3D printing material properties}
Since blood, tissue, and bones have different NIR light properties, we investiagated the properties of different colours and densities of polylactic acid (PLA). There are many different brands that produce PLA, and each has several colours to choose from. We used PLA printing materials from Rankforce \cite{rankforce}.

\begin{figure}[ht]
	\centering
	\includegraphics[width=0.75\columnwidth]{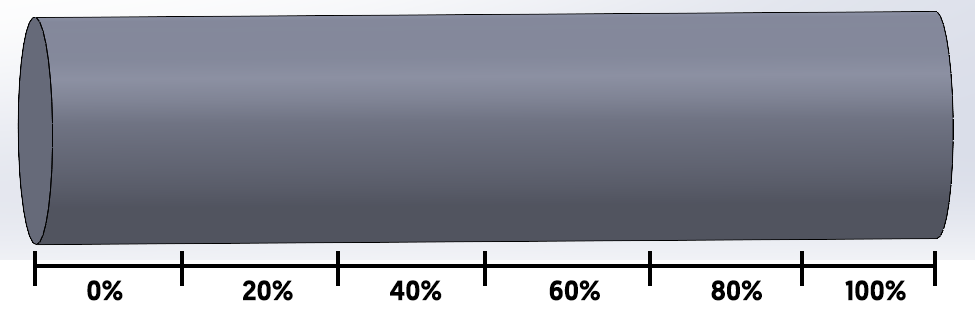}
	\scalebox{-1}[1]{\includegraphics[width=0.75\columnwidth]{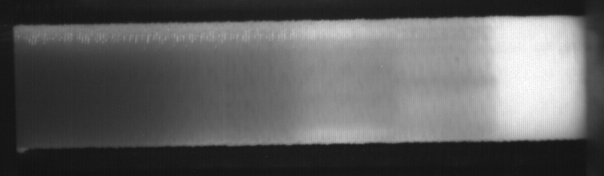}}
	\caption{NIR scan of 3D printed cylinder with varying densities}
	\label{fig6-7}
\end{figure}

We first created a cylinder in Solidworks \cite{solidworks} and using the slicer software Simplify3D \cite{simplify3d} varied the density of the printing material. The 3D printed cylinder was scanned using our fingervein sensor \cite{bram2} and the cylinder and resulting image are shown in Figure \ref{fig6-7}.

\begin{figure}[ht]
	\centering
	\includegraphics[width=\columnwidth]{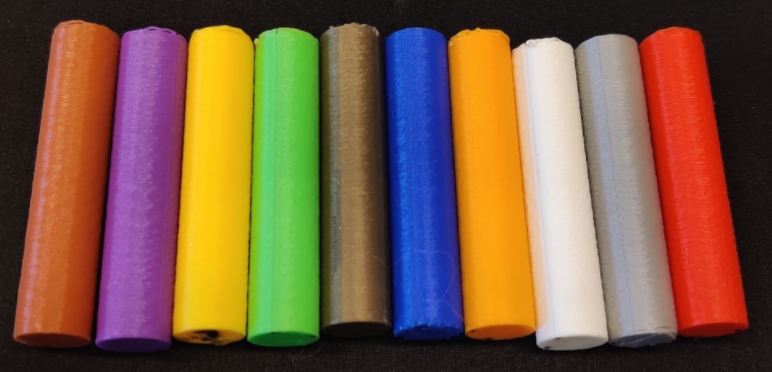}
	\caption{3D printed coloured cylinders}
	\label{fig9}
\end{figure}

In the figure 0\% means the cylinder is hollow, while 100\% means it is solid. It is clear that using a varying printing density it is easy to mimic different absorption from the tissues in the finger.
We also investigated the absorption for 3D printing materials with different colours for various illumination strengths of the scanner and compared it with the absorption of NIR light of a real finger. In Figure \ref{fig9} the set of 3D printed cylinders with different colours is shown, each of them with varying densities as in Figure \ref{fig6-7}. 

\begin{figure*}[ht]
	\centering
	\includegraphics[width=\textwidth]{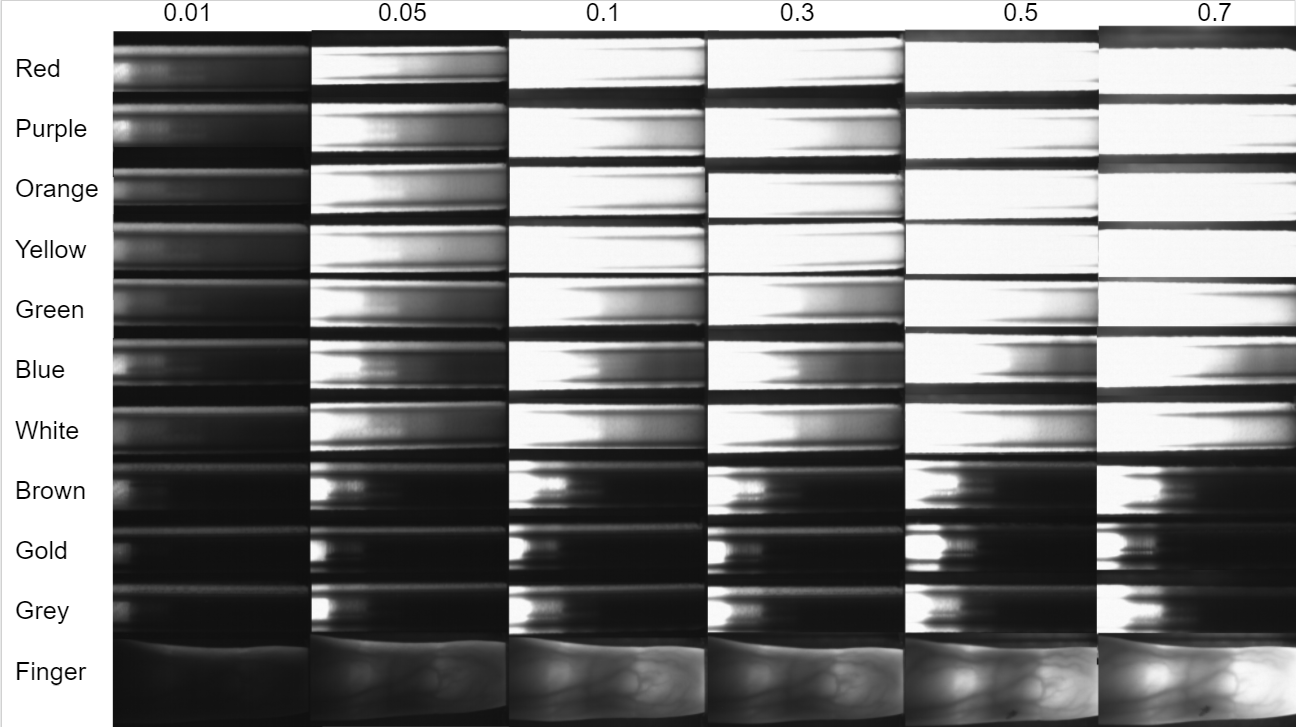}
	\caption{3D printed coloured cylinders in NIR light. The numbers at the top signify illumination strength}
	\label{fig10}
\end{figure*}

The resulting NIR scans of all the cylinders compared to scans of a real finger for varying illumination strength are shown in Figure \ref{fig10}. The optical properties of the Green, Blue and White materials seem closest to those of the real fingers while the Gold and Grey material are closest to the properties of the veins. Therefore, we chose Green for the colour of the bones and soft tissues and Grey for the veins. 

\subsection{Shape of the bones and soft tissue}
\begin{figure}[ht]
	\centering
	\includegraphics[width=0.8\columnwidth]{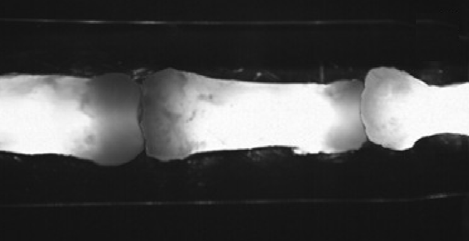}
	\caption{NIR recording of human finger bones}
	\label{figpesi}
\end{figure}

The fingers that are used in finger vein recognition (index, middle and ring fingers) have 3 bones. These bones are thicker near the joints. We recorded a NIR image of human finger bones to investigate their properties in NIR light in \cite{pesi1}, see Figure \ref{figpesi}. From this figure it can be observed that the bone indeed scatters the NIR light in all direction and only absorbs part of the light. At the thicker ends of the bones more light is absorbed, but still quite much of the light passes through and is scattered,

\begin{figure}[ht]
	\centering
	\includegraphics[width=0.8\columnwidth]{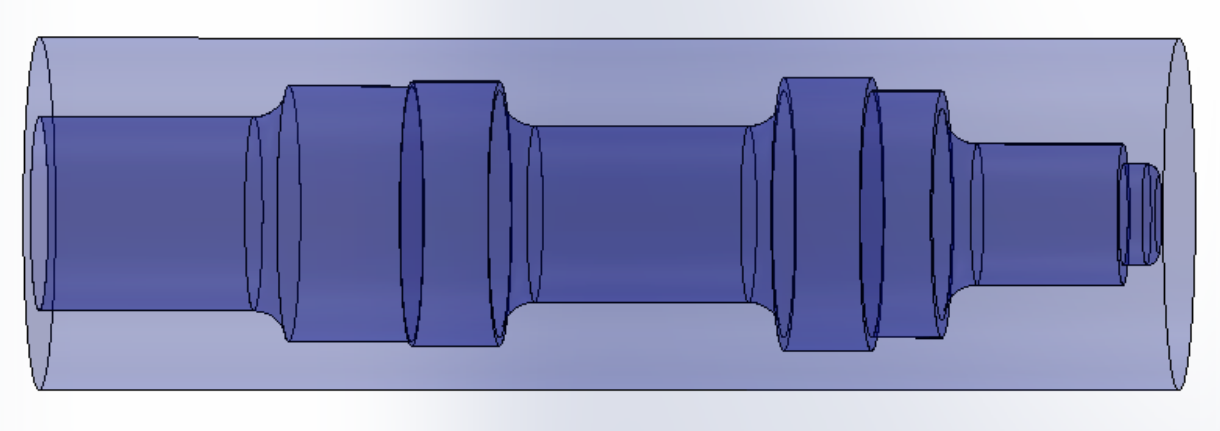}
	\includegraphics[width=0.8\columnwidth]{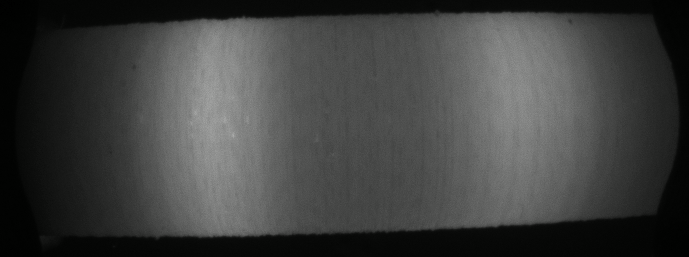}
	\caption{3D model of bones with joints and NIR scan}
	\label{fig13-14}
\end{figure}

We mimicked the basic properties of the bones by creating joints from thicker cylindric parts as illustrated at the top of Figure \ref{fig13-14}. The bone is created using a low density of printing material and the soft tissue that absorbs more NIR light is printed using a higher density. Both use the Green PLA material. As can be observed the soft tissue near the joints is thinner in the model which will result in in a brighter image even though the bones are thicker at the joints and absorb more NIR light, because the absorption of the soft tissue is higher than that of the bone. The result of a NIR recording of the phantom with the bone model and soft tissue is shown at the bottom of Figure \ref{fig13-14}. It shows the same dark and bright bands as the real finger in Figure \ref{fig22}.

\subsection{Defining 3D vein patterns}
\begin{figure}[ht]
	\centering
	\includegraphics[width=0.8\columnwidth]{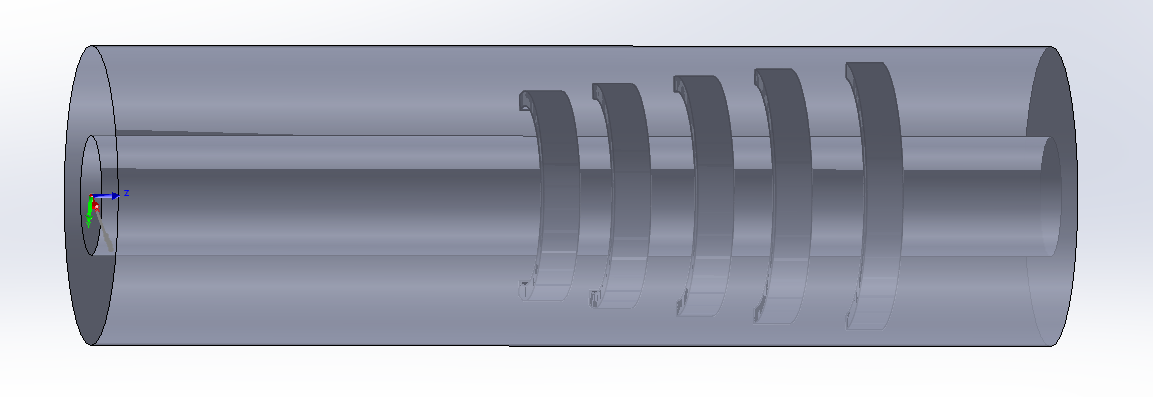}
	\includegraphics[width=0.7\columnwidth]{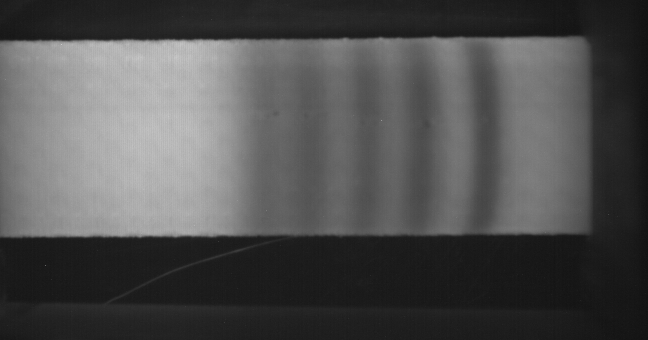}
	\caption{3D vein structures at various depths and NIR image}
	\label{fig11-12}
\end{figure}

We first investigated the impact of the depth of the blood vessels on the image. To this end, we prepared a phantom with vessel like structures with a constant size at various depths from the surface (skin). The 3D model and the NIR image are shown in Figure \ref{fig11-12}. We can see similar effects as in the image of the real finger in Figure \ref{fig22}: the veins closer to the surface, near the joints in the brighter areas are sharper, the other, deeper veins are less sharp.

\begin{figure}[ht]
	\centering
	\includegraphics[width=\columnwidth]{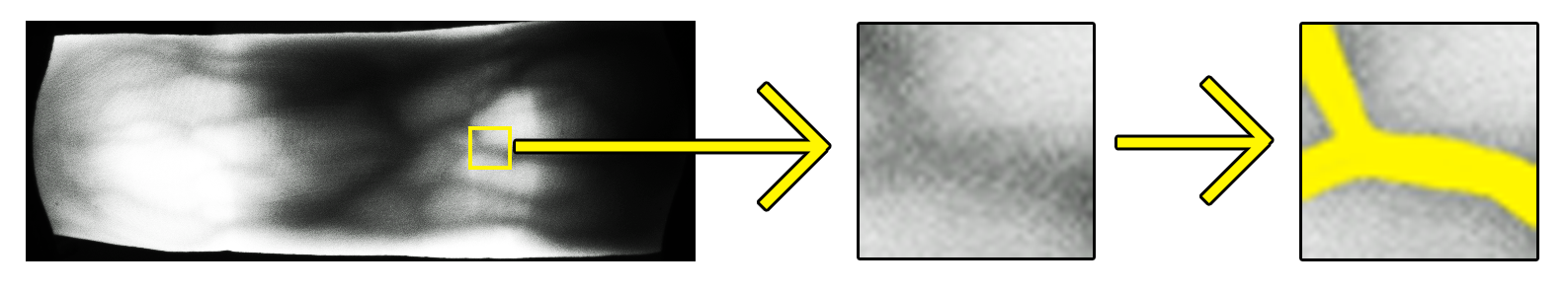}
	\caption{Vein thickness extraction from a real vein image}
	\label{fig15}
\end{figure}

To create realistic vein patterns, we used vein patterns extracted using maximum curvature from real finger vein images. The thickness of the veins is determined from the original finger vein image as shown in Figure \ref{fig15}. 

Now that the skeleton of the veins has been
determined, the veins must be converted into a
3D model. This is done by projection of the 2D pattern on a cylinder with a radius that ensures the veins are at a constant distance to the surface of the finger. This means in our phantom, we do not take varying depths into account, but only the widths of the bloodvessels.
To import the images of the vessels into Solidworks, they need to be transformed into so-called sketches. 
The individual points on the vessels are first converted into contours and then into a DXF file using the Python library exdxf \cite{ezdxf}.
The DXF files can be imported into Solidworks as sketches.

\subsection{Combining into a complete phantom finger}
\begin{figure}[ht]
	\centering
	\includegraphics[width=0.8\columnwidth]{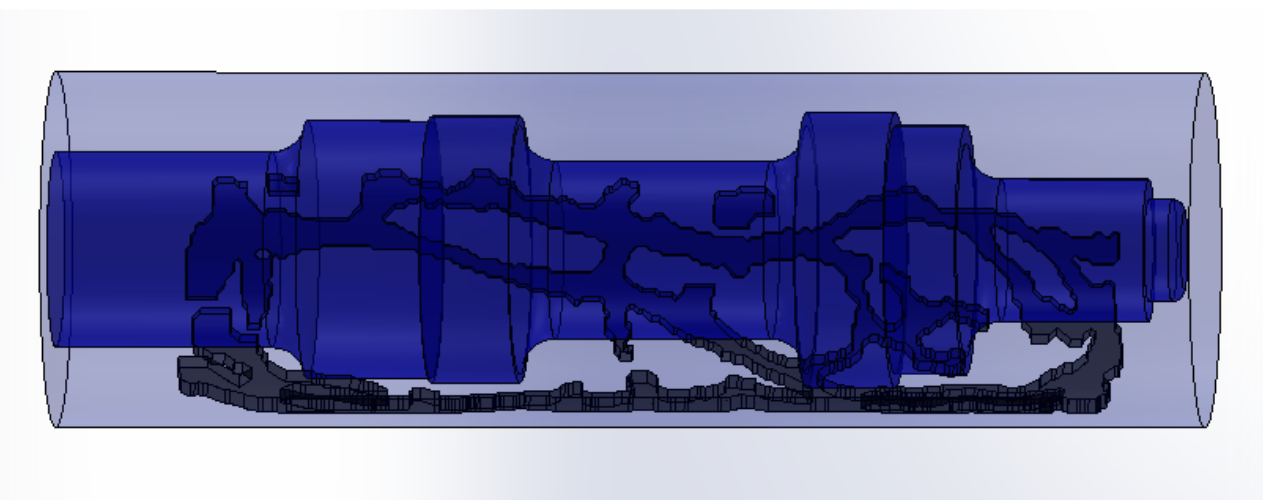}
	\includegraphics[width=0.8\columnwidth]{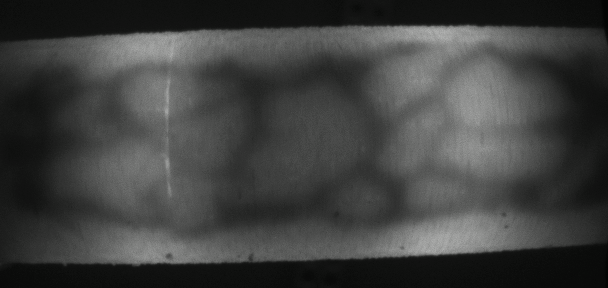}
	\caption{Complete 3D finger phantom model and NIR scan}
	\label{fig19-21}
\end{figure}

The 3D vein pattern DXF file is combined with the 3D model of the bones and soft tissues to form a complete phantom finger. The resulting model is shown in Figure \ref{fig19-21} at the top. The 3D model is printed on a consumer 3D printer with 2 printing heads that can simultaneously print 2 different materials, where Green PLA was used for the bones and soft tissues and Grey for the veins. A NIR image of the resulting phantom is shown at the bottom in Figure \ref{fig19-21}.

By comparing Figures \ref{fig19-21} and \ref{fig22} we can observe that the phantom vein image shows all the main characteristics of the real finger vein image: it has a similar brightness distribution including the dark and bright bands at the joints, the veins also look similar, although the variation in vein widths is less in the phantoms and there is no variation in depth of the veins. Especially very thin veins are missing. This is due to the limitations of the used 3D printer that can only print structures with a thickness of 0.4 mm and larger.

\section{Experiments}
In order to investigate if the phantom finger vein images are similar to real finger vein images for a finger vein recognition system, we set up the following experiment. 

\subsection{Dataset}
First we acquired a small data set of 6 real fingers (of one person). Each finger was recorded 5 times and each time the finger was placed slightly differently on the scanner, i.e. with a different position and rotation. We used rotations up to about 10 degrees. Next we created 6 phantom fingers using a different vein scan of the same 6 fingers and using the procedure described in this paper. The phantom fingers were also scanned 5 times with differen placement and rotation.
We therefore have now two data sets of 30 images in total. For each of the data sets we can create:

\[ N_m = \frac{5\cdot 4}{2}\cdot 6 = 60 \]

mated pairs and

\[ N_{nm}=\frac{5\cdot 5 \cdot 5 }{2} \cdot 6 = 375 \]

non mated pairs.

If we compare phantoms to real fingers, we can create:

\[ N_{mm}=5\cdot 5 \cdot 6 = 150 \]

mixed mated pairs and

\[ N_{mnm}= \frac{10\cdot 10 \cdot 5}{2}\cdot 6 = 1500 \]

non mated pairs.

\subsection{Finger vein comparison}
We used the Miura finger vein comparison approach \cite{miura} with maximum curvature to extract the veins and ICP alignment \cite{pesi2} to compare finger vein pairs. The comparison score is based on the maximum normalised correlation of the binary vein images. We mapped the comparison scores to a range of 0-100, where 0 means no similarity and 100 exactly the same veins.

\subsection{Results}
\begin{figure}[ht]
	\centering
	\includegraphics[width=0.8\columnwidth]{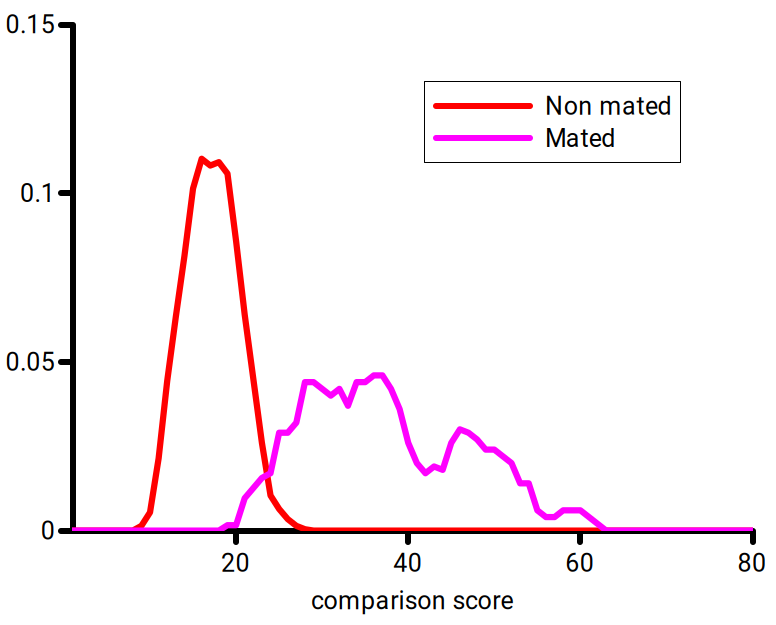}
	\caption{Score histograms of real finger veins}
	\label{figreal}
\end{figure}

In Figure \ref{figreal} the comparison score histograms of mated and non mated finger vein images are shown for the real fingers. We can observe that the non mated scores are all below about 30. The distribution of the mated scores is quite wide, which is caused mainly by the realtively high rotation of the fingers of up to 10 degrees that we allowed. Most mated scores are above 30, though. From experience we know that if the rotation decreases, all mated scores will be above 30 and the overlap between mated and non mated scores disappears. The mated score histograms are not so smooth, beacuse the number of mated pairs is small.

\begin{figure}[ht]
	\centering
	\includegraphics[width=0.8\columnwidth]{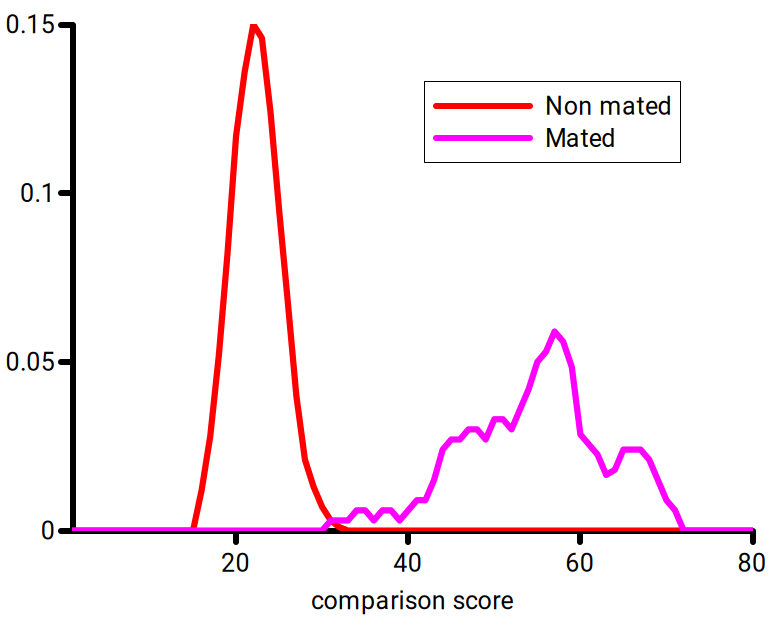}
	\caption{Score histograms of phantom finger veins}
	\label{figphantom}
\end{figure}

In Figure \ref{figphantom} the comparison score histograms of mated and non mated finger vein images are shown for the phantom fingers. Again, we see that all non mated comparison scores are below 30, but the average non mated score is somewhat higher than for the real fingers. We think this is mostly to be attributed to the missing smaller veins in the phantoms. Normally the small veins cause a large diagreement resulting in a lower comparison score. The mated scores are all above 30 and most of them are higher than for the real fingers. We think this may be caused by the missing small veins, that are are harder to detect and for real fingers can reduce the score. Also, in real fingers the musscle tension can create variation of the visibility of the veins, while this is not present in the phantom fingers, of course. Nevertheless, the main behaviour of the comparison scores is similar to that of the real fingers.

\begin{figure}[ht]
	\centering
	\includegraphics[width=0.8\columnwidth]{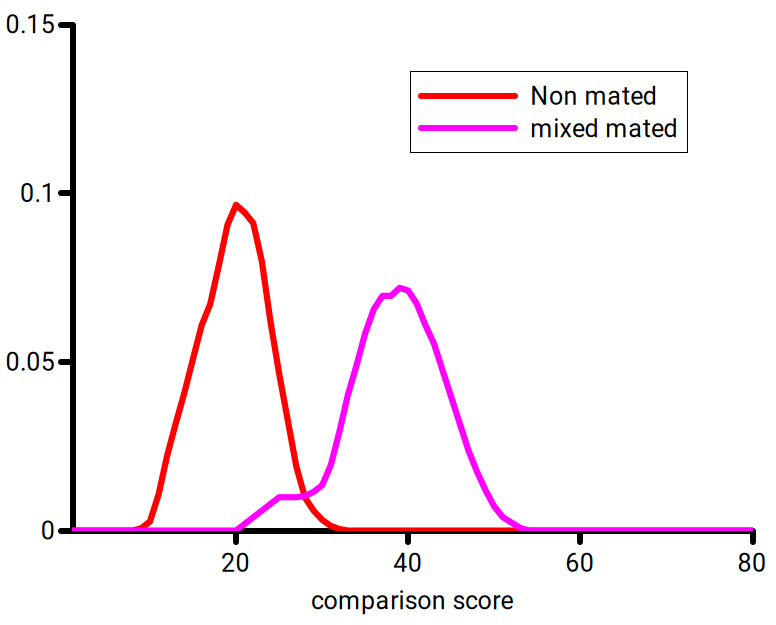}
	\caption{Score histograms of real vs phantom finger veins}
	\label{figmixed}
\end{figure}

Finally, we compared the phantom fingers with the real fingers. The mated scores are now obtained by picking a real finger vein image and a phantom finger vein image that was created using the vein pattern of the real finger. The non mated scores are a combination of all possible non mated scores, i.e. comparisons of real vs real, phantom vs fhantom and real vs phantom. The score histograms for the phantom vs real comparisons are shown in Figure \ref{figmixed}. We can observe that the non mated scores are again all below 30 and the distribution is similar to those of the real and phantom fingers. The mated scores are mostly above 30 again and the distribution is between those of the real and fake fingers. The scores are somewhat higher than those of the real fingers, probably because in one of the images the smaller veins are missing, but lower than those of the phantom fingers, because in one of the images (the real finger) the smaller veins are present and lead to lower similarity.

\subsection{Presentation Attack}
The fact that most of the mated comparisons between phantom and real fingers result in scores above 30 and in the same range as real finger vein comparisons, means that using the phantom, it is possible to spoof the finger vein sensor. The phantoms exhibit the same properties as real fingers, i.e. they show the same grey level distribution and vein patterns as real fingers in NIR illumination. Of course visual inspection of the phantoms would immediately reveal that they are cylindric and have no natural shape, not to mention that they are green!

\section{Conclusion}
The aim of the presented research was to investigate if using 3D printing techniques it is possible to create finger vein phantoms that mimic the behaviour of real fingers in NIR illumination closely. We found that the absorption and scattering behaviour of bones, soft tissues and veins can be approximated by using the proper 3D printing materials and varying the density of the printing material in the printing process. For bones and soft tissues we chose green PLA where for bones we used less dense and for soft tissues denser printing, because bones absorb NIR less than soft tissues. For the veins we used grey PLA and used solid printing. This resulted in very high absorption of NIR red light like for real veins. We also took into account the shape of the bones that are thicker near the joints and were able to replicate the brighter bands around the joints in finger vein images, caused by a thinner layer of soft tissues at that location. To create realistic vein patterns, we used 2D scans of real vein patterns and transformed them into 3D and combined these with a cylindric model of the bones and soft tissues. The complete phantoms could be 3D printed using a 3D printer with two heads to print two materials (green and grey) simultaneously. 
It turned out that the thus created phantom fingers produced convincing finger vein images when scanned using NIR light. We collected a small dataset with 6 fingers and showed that the score distributions of mated and non mated pairs when compared using Miura's finger vein recognition method show similar behaviour and that phantoms created using the vein pattern of a finger can be used successfully for Presentation Attacks for that finger. 

\bigskip
We use the finger vein phantoms currently for improving the design of finger vein scanner devices and for investigating the behaviour of methods for finger vein pattern extraction and recognition. The current vein phantoms don't have very thin veins due to limitations of the 3D printer, which has a nozzle of 0.4 mm. We intend to replace it with a finer nozzle of 0.1 mm. Furthermore, the finger phantoms can be made more realistic by using an actual finger shape instead of a cylinder. Also we plan to use synthesized finger vein patterns instead of scanned real finger vein patterns and vary the depth of the veins in the patterns to further approximate real fingers.

\begin{figure}[b]
	\centering
	\mbox{\begin{minipage}{5cm}\vspace{2.0cm}\end{minipage}}
\end{figure}

\bibliography{Spreeuwers_fingerphantoms}
\bibliographystyle{IEEEtran}

\end{document}